\documentclass[letterpaper, 10 pt, conference]{ieeeconf}  

\IEEEoverridecommandlockouts                              

\input{dependencies}

\newacronym{mdp}{MDP}{Markov Decision Process}
\newacronym{rl}{RL}{Reinforcement Learning}
\newacronym{dl}{DL}{Deep Learning}
\newacronym{mav}{MAV}{Micro Aerial Vehicle}
\newacronym{gpu}{GPU}{Graphics Processing Unit}
\newacronym{ml}{ML}{Machine Learning}
\newacronym{dnn}{DNN}{Deep Neural Network}
\newacronym{trpo}{TRPO}{Trust Region Policy Optimization}
\newacronym{ddpg}{DDPG}{Deep Deterministic Policy Gradient}
\newacronym{gae}{GAE}{Generalized Advantage Estimation}
\newacronym{ppo}{PPO}{Proximal Policy Optimization}
\newacronym{gps}{GPS}{Guided Policy Search}
\newacronym{cnn}{CNN}{Convolutional Neural Network}
\newacronym{rnn}{RNN}{Recurrent Neural Network}
\newacronym{relu}{ReLU}{Rectified Linear Unit}
\newacronym{tqc}{TQC}{Trucated Quantile Critics}
\newacronym{hpc}{HPC}{High Performance Computing}
\newacronym{fcu}{FCU}{Flight Controller Unit}
\newacronym{fma}{FMA}{Fused Multiply-Accumulate}
\newacronym{nvcc}{nvcc}{Nvidia CUDA Compiler}
\newacronym{stl}{STL}{Standard Template Library}
\newacronym{td3}{TD3}{Twin Delayed Deep Deterministic policy gradient}
\newacronym{vmt}{VMT}{Virtual Method Table}
\newacronym{rpm}{RPM}{Revolutions Per Minute}
\newacronym{uav}{UAV}{Unmanned Aerial Vehicle}
\newacronym{rk4}{RK4}{$4^{\text{th}}$ order Runge-Kutta}
\newacronym{gemm}{GEMM}{GEneral Matrix Multiply}
\newacronym{sfinae}{SFINAE}{Substitution Failure Is Not An Error}
\newacronym{ram}{RAM}{Random Access Memory}
\newacronym{iot}{IoT}{Internet of Things}
\newacronym{gcc}{GCC}{GNU Compiler Collection}
\newacronym{cots}{COTS}{Commercial Off-The-Shelf}
\newacronym{mems}{MEMS}{Microelectromechanical Systems}
\newacronym{vtol}{VTOL}{Vertical Take-Off and Landing}
\newacronym{pwm}{PWM}{Pulse-Width Modulation}
\newacronym{ui}{UI}{User Interface}
\newacronym{ctbr}{CTBR}{Collective Thrust and Body Rates}
\newacronym{srt}{SRT}{Single Rotor Thrusts}
\newacronym{esc}{ESC}{Electronic Speed Controller}
\newacronym{sim2real}{Sim2Real}{Simulation-to-Reality}
\newacronym{rmse}{RMSE}{Root-Mean-Square Error}
\newacronym{bpo}{BPO}{Bellman's Principle of Optimality}
\newacronym{indi}{INDI}{Incremental Non-Linear Dynamic Inversion}
\newacronym{pid}{PID}{Proportional–Integral–Derivative}
\newacronym{aac}{AAC}{Asymmetric Actor-Critic}
\newacronym{arpl}{ARPL}{Agile Robotics and Perception Lab}
\newacronym{map}{MAP}{Maximum A Posteriori}
\newacronym{mpc}{MPC}{Model Predictive Control}
\newacronym{mle}{MLE}{Maximum Likelihood Estimation}
\newacronym{ema}{EMA}{Exponential Moving Average}
\newacronym{wlog}{w.l.o.g}{Without loss of generality}
\newacronym{ode}{ODE}{Ordinary Differential Equation}

\definecolor{known}{RGB}{0,123,255}    
\definecolor{unknown}{RGB}{220,20,60} 
\definecolor{observable}{RGB}{218,165,32} 
\DeclareMathOperator*{\argmax}{argmax}
\DeclareMathOperator*{\argmin}{argmin}
\DeclareMathOperator*{\diag}{diag}
\newcommand{\grayrow}{\rowcolor{gray!10}}
\newcommand{\captionvspace}{\vspace{-15pt}}

\title{\LARGE \bf
Data-Driven System Identification of Quadrotors\\ Subject to Motor Delays
}

\author{Jonas Eschmann$^{1,2}$, Dario Albani$^{2}$, and Giuseppe Loianno$^{1}$
\thanks{
$^1$The authors are with the New York University, Tandon School of Engineering, Brooklyn, NY 11201, USA. {\tt\footnotesize email: \{jonas.eschmann, loiannog\}@nyu.edu}.}
\thanks{$^2$The authors are with the Autonomous Robotics Research Center, Technology Innovation Institute, Abu Dhabi, UAE. {\tt\footnotesize email: \{jonas.eschmann, dario.albani\}@tii.ae}.}
\thanks{ This work was supported by the Technology Innovation Institute, the NSF CAREER Award 2145277, and the DARPA YFA Grant D22AP00156-00. Giuseppe Loianno serves as consultant for the Technology Innovation Institute. This arrangement has been reviewed and approved by the New York University in accordance with its policy on objectivity in research.}
}

\begin{document}

\maketitle
\thispagestyle{empty}
\pagestyle{empty}

\begin{abstract}
Recently non-linear control methods like \gls*{mpc} and \gls*{rl} have attracted increased interest in the quadrotor control community. In contrast to classic control methods like cascaded PID controllers, \gls*{mpc} and \gls*{rl} heavily rely on an accurate model of the system dynamics. The process of quadrotor system identification is notoriously tedious and is often pursued with additional equipment like a thrust stand. Furthermore, low-level details like motor delays which are crucial for accurate end-to-end control are often neglected. In this work, we introduce a data-driven method to identify a quadrotor's inertia parameters, thrust curves, torque coefficients, and first-order motor delay purely based on proprioceptive data. The estimation of the motor delay is particularly challenging as usually, the RPMs can not be measured. We derive a \gls*{map}-based method to estimate the latent time constant. Our approach only requires about a minute of flying data that can be collected without any additional equipment and usually consists of three simple maneuvers. Experimental results demonstrate the ability of our method to accurately recover the parameters of multiple quadrotors. It also facilitates the deployment of \gls*{rl}-based, end-to-end quadrotor control of a large quadrotor under harsh, outdoor conditions.
\end{abstract}

\section*{Supplementary Material}
\noindent \textbf{Video}: \href{https://youtu.be/G3WGthRx2KE}{https://youtu.be/G3WGthRx2KE}

\noindent \textbf{Code}: \href{https://github.com/arplaboratory/data-driven-system-identification}{\url{https://github.com/arplaboratory/data-driven-system-identification}}

\noindent \textbf{Project page}: \href{https://sysid.tools}{https://sysid.tools}

\section{Introduction}
Aerial vehicles, such as \gls*{mav}, have proven to be a versatile platform for numerous real-world problems like search and rescue, infrastructure inspection, or package delivery. While the applications have been evolving, the dominating control stack in terms of real-world deployment (e.g., as part of the PX4 Autopilot) has not changed at the same pace. To this date, this cascade of \gls*{pid} controllers has proven itself by market demand but from a theoretical perspective, it is not satisfying as it may leave performance at the table (as discussed e.g., in \cite{eschmann2023learning}). Moreover, from an end-user perspective, tuning the \gls*{pid} controller parameters is a notoriously daunting task. 
\begin{table}[tb]
\centering
\def\arraystretch{1.2}
\begin{adjustbox}{width=1.0\columnwidth}
\setlength{\tabcolsep}{1mm}
\begin{tabular}{c l}
\hline
Variable & Description \\
\hline
\grayrow
$\mathbf{p}$ / $\mathbf{v}$ & Global position / velocity \\
$\mathbf{q}$ / $\mathbf{R}(\mathbf{q})$ & Orientation: quaternion / rotation matrix \\
\grayrow
$f_i$ & Thrust produced by motor $i$ \\
$\boldsymbol{\omega}_m$ / $\textcolor{observable}{\boldsymbol{\omega}_{sp}}$ & Motor RPMs: state / setpoints \\
\grayrow
$\textcolor{observable}{\boldsymbol{\omega}_b}$ / $\textcolor{observable}{\mathbf{o}_{acc}}$  & Angular rate / accelerometer meas. (body frame) \\
$\textcolor{known}{m}$ / $\textcolor{known}{\mathbf{g}}$ & Mass / gravity \\
\grayrow
$\textcolor{known}{\mathbf{r}_{p_i}}$ / $\textcolor{known}{\mathbf{r}_{f_i}}$ / $\textcolor{known}{\mathbf{r}_{\tau_i}}$  & Motor $i$: position / force / torque \\
$\textcolor{unknown}{\textbf{J}}$ / $\textcolor{unknown}{T_m}$ & Inertia matrix / Motor delay time constant \\
\grayrow
$\textcolor{unknown}{K_{\tau_{i}}}$ / $\textcolor{unknown}{K_{f_{ij}}}$ & Motor $i$: torque coeff. / thrust coeff. (exponent $j$) \\
\hline
\multirow{2}{*}{
\tikz\draw[fill=.] (0,0) circle (.7ex); }
& Defined in terms of
\textcolor{observable}{\tikz\draw[fill=.] (0,0) circle (.7ex);}
\textcolor{known}{\tikz\draw[fill=.] (0,0) circle (.7ex);}
\textcolor{unknown}{\tikz\draw[fill=.] (0,0) circle (.7ex);} \\
& or not relevant to the system identification \\
\textcolor{observable}{\tikz\draw[fill=.] (0,0) circle (.7ex);}
\textcolor{known}{\tikz\draw[fill=.] (0,0) circle (.7ex);}
\textcolor{unknown}{\tikz\draw[fill=.] (0,0) circle (.7ex);}
& \textcolor{observable}{observable (proprioceptive)}, \textcolor{known}{known (a priori)}, \textcolor{unknown}{unknown} \\
\hline
\end{tabular}
\end{adjustbox}
\caption{Notation.}
\label{table:notation}
\vspace{-10pt}
\end{table}
Modern alternatives to this control stack like \gls*{mpc} or end-to-end \gls*{rl} bear promising theoretical properties and have shown good performance under controlled conditions, but to date did not see large-scale adoption. One limitation and major difference of \gls*{mpc} and \gls*{rl} from the \gls*{pid} controllers is that they require an accurate model of the system dynamics. Theoretically, \gls*{rl}-based trial-and-error learning could be directly applied to the system but in practice this is not feasible due to the sample complexity. This is particularly true in the case of quadrotors, due to the wear and tear caused by the required random exploration (i.e., crashes).

Hence, for widespread adoption of these non-linear control approaches on a diverse set of platforms, a simple and robust system identification method is required. It should decrease the burden that comes with the system identification process to a point where it is lower than the burden of \gls*{pid} tuning. In this work, we tackle this problem and propose a novel method for quadrotor system identification in form of the following major contributions
\begin{enumerate}
\item \textbf{Data-driven system identification method} that includes motor delays and only relies on proprioceptive observations (accelerometer, gyroscope, and motor commands) and simple measurements (mass, rotor-positions).
\item \textbf{Comparative study} of identified parameters on a commercially available aerial robot with parameters identified in prior works.
\item \textbf{Real-world demonstration} of an end-to-end \gls*{rl} policy deployed on a large quadrotor, trained in simulation, and based on the parameters identified from only \SI{73}{\second} of flight data.
\item \textbf{Open source} release of the implementation of our method including an example for easy reproduction of all the system identification results presented in this work.
\item \textbf{System identification software} for even simpler identification of quadrotor parameters. We provide an open-source, interactive and web-based software for researchers and practitioners to identify quadrotor parameters by uploading their PX4 ULog or Crazyflie logfiles which are popular logging formats in research and industry. 
\end{enumerate}
\section{Related Works}

Numerous works have been published, usually studying the parameter identification of a particular system instead of proposing a general method. Several works~\cite{foerster2015systemidentificationofthecrazyflie, luis2016design, greiff2017modelling, nguyen2023crazyflie, rich2012model} use white-box system-identification methods where the system is tested using additional equipment like a thrust-stand or even disassembled. Other works~\cite{alabsi2019real,alabsi2019quadrotor,cho2019system,wei2014frequency,adiprawita2007automated,ivler2019multirotor,yuksek2020system,babcock2023system} can be considered data-driven but they all use proprietary software, obstructing reproducibility. \cite{galliker2021data,wuest2019online,burri2018mlesysid} use data-driven system identification. However, they do not take into account motor delays and they use black-box optimization instead of closed-form solutions where possible. Additionally, the most related work \cite{galliker2021data} focuses on the aerodynamic effects of airfoils. Our work instead decomposes the problem, starting from the gray-box model, and finds closed-form solutions by casting the parameter identification as \gls*{mle}, naturally giving rise to a convex optimization problem formulation. For the estimation of the motor delays we apply \gls*{map} estimation and find that we can robustly find a global optimum.

\section{Methodology}
\label{sec:methodology}
In this section, we outline the methodology behind our work and start by defining the standard dynamics of a quadrotor subject to motor delay (e.g., \cite{kaufmann2023champion}):
\begin{align}
\dot{\mathbf{p}} = & \; \mathbf{v}, \\
\dot{\mathbf{q}} = & \; \mathbf{q} \circ \left[0 \;\; \textcolor{observable}{\boldsymbol{\omega}_b}/ 2\right]^\top,\\
\dot{\mathbf{v}} = & \; \frac{1}{\textcolor{known}{m}}\mathbf{\mathbf{R}}(\mathbf{q}) \left(\sum_{i=1}^4 \textcolor{known}{\mathbf{r}_{f_i}} f_i\right) + \textcolor{known}{\mathbf{g}}, \\
\dot{\mathbf{v}} = & \; \mathbf{R}\left(\mathbf{q}\right)\dot{\textbf{v}}_b, \\
\dot{\textbf{v}}_b = & \; \textcolor{observable}{\mathbf{o}_{\text{acc}}} + \textbf{R}\left(\mathbf{q}\right)^{-1}\textcolor{known}{\textbf{g}}, \label{eq:v_b_dot_definition}\\
f_i = & \; \sum_{j=0}^{2} \textcolor{unknown}{K_{f_{ij}}} \omega_{m_i}^j, \label{eq:thrust_curve}\\
\dot{\boldsymbol{\omega}}_b = & \; \mathbf{\textcolor{unknown}{J}}^{-1}\left(\boldsymbol{\tau} + \left(\textcolor{unknown}{\mathbf{J}}\textcolor{observable}{\boldsymbol{\omega}_b}\right) \times  \textcolor{observable}{\boldsymbol{\omega}_b}\right), \label{eq:angular_dynamics}\\
\dot{\boldsymbol{\omega}}_b = & \; \frac{\text{d}}{\text{dt}} \textcolor{observable}{\boldsymbol{\omega}_b}, \label{eq:angular_acceleration}\\
\boldsymbol{\tau} = & \; \sum_{i=1}^4 \left(\textcolor{known}{\mathbf{r}_{p_i}} \times \textcolor{known}{\mathbf{r}_{f_i}}\right)f_i + \textcolor{known}{\mathbf{r}_{\tau_i}}\textcolor{unknown}{K_{\tau_i}} f_i,  \label{eq:torque}\\
\dot{\boldsymbol{\omega}}_m = & \; \textcolor{unknown}{T_m}^{-1} \left(\textcolor{observable}{\boldsymbol{\omega}_{sp}} - \boldsymbol{\omega}_m\right). 
\label{eq:motor_dynamics}\\
\end{align}
Note that eq.~\eqref{eq:motor_dynamics} describes the motors as decoupled first-order systems subject to the time-constant $\textcolor{unknown}{T_m}$. 

Now we can solve for the thrust curve with the goal to express it in terms of \textcolor{known}{known} or \textcolor{observable}{observable} variables
\label{sec:motor_model_estimation}
\begin{align}
\mathbf{R}\left(\mathbf{q}\right)\dot{\textbf{v}}_b = & \; \frac{1}{\textcolor{known}{m}}\mathbf{\mathbf{R}}(\mathbf{q}) \left(\sum_{i=1}^4 \textcolor{known}{\mathbf{r}_{f_i}} f_i\right) + \textcolor{known}{\mathbf{g}}, \\
\dot{\textbf{v}}_b = & \; \frac{1}{\textcolor{known}{m}}\left(\sum_{i=1}^4 \textcolor{known}{\mathbf{r}_{f_i}} f_i\right) + \mathbf{\mathbf{R}}(\mathbf{q})^{-1} \textcolor{known}{\mathbf{g}}, \\
\dot{\textbf{v}}_b - \mathbf{\mathbf{R}}(\mathbf{q})^{-1} \textcolor{known}{\mathbf{g}} = & \; \frac{1}{\textcolor{known}{m}}\sum_{i=1}^4 \textcolor{known}{\mathbf{r}_{f_i}} f_i.\\
\end{align}
Inserting eq.~\eqref{eq:v_b_dot_definition} and eq.~\eqref{eq:thrust_curve}:
\begin{align}
\textcolor{observable}{\mathbf{o}_{\text{acc}}} + \textbf{R}\left(\mathbf{q}\right)^{-1}\textcolor{known}{\textbf{g}} - \mathbf{\mathbf{R}}(\mathbf{q})^{-1} \textcolor{known}{\mathbf{g}}  = & \; \frac{1}{\textcolor{known}{m}} \sum_{i=1}^4 \textcolor{known}{\mathbf{r}_{f_i}} \sum_{j=0}^{2} \textcolor{unknown}{K_{f_{ij}}} \omega_{m_i}^j, \\
\textcolor{known}{m} \textcolor{observable}{\mathbf{o}_{\text{acc}}} = & \; \sum_{i=1}^4 \sum_{j=0}^{2} \textcolor{known}{\textcolor{unknown}{K_{f_{ij}}} \mathbf{r}_{f_i}}  \omega_{m_i}^j.\\
\end{align}
This is linear in the unknowns $\textcolor{unknown}{K_{f_{ij}}}$. \\
\textbf{Goal}: writing this in the form $\mathbf{A}\mathbf{x} = \mathbf{b}$ with
\begin{align}
\mathbf{b} := & \textcolor{known}{m}\textcolor{observable}{\mathbf{o}_\text{acc}}, \\
\mathbf{x} := & \left[\textcolor{unknown}{K_{f_{10}}}, \textcolor{unknown}{K_{f_{11}}}, \textcolor{unknown}{K_{f_{12}}}, \ldots, \textcolor{unknown}{K_{f_{40}}}, \textcolor{unknown}{K_{f_{41}}}, \textcolor{unknown}{K_{f_{42}}}\right]^\top, \label{eqn:single_sample_parameters} \\
\begin{split}
\mathbf{A} := & [\textcolor{known}{\textbf{r}_{f_0}}\omega_{m_0}^0, \textcolor{known}{\textbf{r}_{f_0}}\omega_{m_0}^1, \textcolor{known}{\textbf{r}_{f_0}}\omega_{m_0}^2, \ldots \\
& \qquad \qquad \ldots, \textcolor{known}{\textbf{r}_{f_4}}\omega_{m_4}^0, \textcolor{known}{\textbf{r}_{f_4}}\omega_{m_4}^1, \textcolor{known}{\textbf{r}_{f_4}}\omega_{m_4}^2]. \label{eqn:single_sample_data_matrix}
\end{split}
\end{align}
The only missing part for the linear equation to be fully specified are the rotor speeds $\omega_{m_i} \forall i \in 1 \ldots 4$. We can not directly observe $\omega_{m_i}$ (which would also allow solving for the time constant in closed-form, like the other parameters), but we can solve for it by solving the first-order \gls*{ode} assuming a given time-constant $\textcolor{unknown}{T_m}$ and the \gls*{rpm} setpoints $\textcolor{observable}{\boldsymbol{\omega}_{sp}}$ (time-dependent). Since the dynamics of the motors are independent we can solve them per motor as:
\begin{align}
\dot{\omega}_{m_i}\left(t\right) = & \; \textcolor{unknown}{T_m}^{-1} \left(\textcolor{observable}{\omega_{{sp}_i}}\left(t\right) - \omega_{m_i}\left(t\right)\right). \\
\end{align}
Ansatz for solving the integral: multiply by $\mu\left(t\right)= e^{\frac{t}{\textcolor{unknown}{T_m}}}$
\begin{align}
\mu\left(t\right)\dot{\omega}_{m_i}\left(t\right) + \mu\left(t\right)\textcolor{unknown}{T_m}^{-1} &  \omega_{m_i}\left(t\right) =  \mu\left(t\right)\textcolor{unknown}{T_m}^{-1} \textcolor{observable}{\omega_{{sp}_i}}\left(t\right), \\
\frac{\text{d}}{\text{dt}}\left(\mu\left(t\right)\omega_{m_i}\left(t\right)\right) = & \; \mu\left(t\right)\textcolor{unknown}{T_m}^{-1} \textcolor{observable}{\omega_{{sp}_i}}\left(t\right), \\
\int_0^t \frac{\text{d}}{\text{d}s}\left(\mu\left(s\right)\omega_{m_i}\left(s\right)\right) \text{d}s = & \; \int_0^t \mu\left(s\right)\textcolor{unknown}{T_m}^{-1} \textcolor{observable}{\omega_{{sp}_i}}\left(s\right) \text{d}s, \\
\mu\left(t\right)\omega_{m_i}\left(t\right) = & \; \int_0^t \mu\left(s\right)\textcolor{unknown}{T_m}^{-1} \textcolor{observable}{\omega_{{sp}_i}}\left(s\right) \text{d}s. \\
\end{align}
Then $\omega_{m_i}\left(t\right)$ can be expressed in terms of an integral transformation
\begin{align}
\omega_{m_i}\left(t\right) = & \; e^{-\frac{t}{\textcolor{unknown}{T_m}}} \int_0^t e^{\frac{s}{\textcolor{unknown}{T_m}}}\textcolor{unknown}{T_m}^{-1} \textcolor{observable}{\omega_{{sp}_i}}\left(s\right) \text{d}s \\
= & \; \textcolor{unknown}{T_m}^{-1}\int_0^t e^{\frac{s-t}{\textcolor{unknown}{T_m}}} \textcolor{observable}{\omega_{{sp}_i}}\left(s\right) \text{d}s \\
= & \; \textcolor{unknown}{T_m}^{-1}\int_0^t e^{\frac{-\left(t-s\right)}{\textcolor{unknown}{T_m}}} \textcolor{observable}{\omega_{{sp}_i}}\left(s\right) \text{d}s. \\
\end{align}
This form can also be trivially reformulated in terms of a convolution but for our application, we are interested in a discrete-time solution.
To find a discrete-time solution, first we find the continuous time solution for a time step $\Delta t$
\begin{align}
\omega_{m_i}\left(0\right) = & \; 0, \\
\omega_{m_i}\left(t + \Delta t\right) = & \; \textcolor{unknown}{T_m}^{-1}\int_0^{t+\Delta t} e^{\frac{-\left(t + \Delta t-s\right)}{\textcolor{unknown}{T_m}}} \textcolor{observable}{\omega_{{sp}_i}}\left(s\right) \text{d}s \\
= & \; \textcolor{unknown}{T_m}^{-1}\int_0^{t} e^{\frac{-\left(t + \Delta t-s\right)}{\textcolor{unknown}{T_m}}} \textcolor{observable}{\omega_{{sp}_i}}\left(s\right) \text{d}s\\
& + \textcolor{unknown}{T_m}^{-1}\int_t^{t+\Delta t} e^{\frac{-\left(t + \Delta t-s\right)}{\textcolor{unknown}{T_m}}} \textcolor{observable}{\omega_{{sp}_i}}\left(s\right) \text{d}s \\
= & \; e^{\frac{-\Delta t}{\textcolor{unknown}{T_m}}}\overbrace{\textcolor{unknown}{T_m}^{-1}\int_0^{t} e^{\frac{-\left(t-s\right)}{\textcolor{unknown}{T_m}}} \textcolor{observable}{\omega_{{sp}_i}}\left(s\right) \text{d}s}^{\omega_{m_i}\left(t\right)} \label{eq:motor_model_recursive_solution}\\
& + \textcolor{unknown}{T_m}^{-1}\int_t^{t+\Delta t} e^{\frac{-\left(t + \Delta t-s\right)}{\textcolor{unknown}{T_m}}} \textcolor{observable}{\omega_{{sp}_i}}\left(s\right) \text{d}s. \\
\end{align}
In the following, we take advantage of the recursive structure in eq.~\eqref{eq:motor_model_recursive_solution}. Now we can move to discrete-time control inputs $\textcolor{observable}{\omega_{{sp}_i}}$ by assuming $g\left(t\right) = \textcolor{observable}{\omega_{{sp}_i}}\left(t\right)$ is constant for $\Delta t$
\begin{align}
\omega_{m_i}\left(t + \Delta t\right) = & \; e^{\frac{-\Delta t}{\textcolor{unknown}{T_m}}}\omega_{m_i}\left(t\right)\\
& + \textcolor{unknown}{T_m}^{-1}\textcolor{observable}{\omega_{{sp}_i}}\left(t\right)\int_t^{t+\Delta t} e^{\frac{-\left(t + \Delta t-s\right)}{\textcolor{unknown}{T_m}}} \text{d}s \\
= & \; e^{\frac{-\Delta t}{\textcolor{unknown}{T_m}}}\omega_{m_i}\left(t\right)\\
& + \textcolor{unknown}{T_m}^{-1}\textcolor{observable}{\omega_{{sp}_i}}\left(t\right) e^{\frac{-\left(t + \Delta t\right)}{\textcolor{unknown}{T_m}}} \underbrace{\int_t^{t+\Delta t} e^{\frac{s}{\textcolor{unknown}{T_m}}} \text{d}s}_{\left[\textcolor{unknown}{T_m} e^\frac{s}{\textcolor{unknown}{T_m}}\right]_t^{t + \Delta t}} \\
= & \; e^{\frac{-\Delta t}{\textcolor{unknown}{T_m}}}\omega_{m_i}\left(t\right) + \textcolor{observable}{\omega_{{sp}_i}}\left(t\right) e^{\frac{-\left(t + \Delta t\right)}{\textcolor{unknown}{T_m}}} \left[e^\frac{s}{\textcolor{unknown}{T_m}}\right]_t^{t + \Delta t} \\
= & \; e^{\frac{-\Delta t}{\textcolor{unknown}{T_m}}}\omega_{m_i}\left(t\right) + \textcolor{observable}{\omega_{{sp}_i}}\left(t\right) \left[1 -e^{\frac{-\Delta t}{\textcolor{unknown}{T_m}}}\right]. \\
\end{align}
We recognize the form of an \gls*{ema} which provides a computationally efficient way to infer the motor speeds based on motor speed setpoints
\begin{align}
\alpha := & \; e^{\frac{-\Delta t}{\textcolor{unknown}{T_m}}},  \\
\omega_{m_i}\left(t + \Delta t\right) = & \; \alpha \omega_{m_i}\left(t\right) + \textcolor{observable}{\omega_{{sp}_i}}\left(t\right) \left[1-\alpha\right]. \\
\end{align}
Using the motor model and given a $\textcolor{unknown}{T_m}$, we can retrieve the motor speeds at any point in time. This fully specifies $\mathbf{A}$ (cf. eq.~\eqref{eqn:single_sample_data_matrix}) and hence the linear equation
$\textbf{A}_k \textbf{x}_k = \textbf{b}_k$
where each sample $k$ gives rise to $3$ linear equations.

To find $\textcolor{unknown}{T_m}$, we define the conditional likelihood of our dataset $\mathcal{D}$ by using an isotropic Gaussian observer model
\begin{align}
p\left(\mathcal{D} | \textcolor{unknown}{T_m}\right) = & \; \prod_{k=1}^{|\mathcal{D}|} p\left(\mathbf{b}_k | \mathbf{A}_k\right), \\
p\left(\mathbf{b}_k | \mathbf{A}_k\right) = & \; \mathcal{N}\left(\mathbf{b}_k; \mathbf{A}_k \mathbf{x}, \mathbf{I}\right), \\
p\left(\textcolor{unknown}{T_m} | \mathcal{D}\right) \propto & \; p\left(\mathcal{D} | \textcolor{unknown}{T_m}\right) p\left(\textcolor{unknown}{T_m}\right) \\
= & \; \prod_{k=1}^{|\mathcal{D}|} p\left(\mathbf{b}_k | \mathbf{A}_k\right) p\left(\textcolor{unknown}{T_m}\right).\\
\end{align}
We want to find the \gls*{map} estimate of $\textcolor{unknown}{T_m}$
\begin{align}
 & \underset{\textcolor{unknown}{T_m}}{\argmax} \; p\left(\textcolor{unknown}{T_m} | \mathcal{D}\right) = \underset{\textcolor{unknown}{T_m}}{\argmax} \; \log p\left(\textcolor{unknown}{T_m} | \mathcal{D}\right)  \\
& = \; \underset{\textcolor{unknown}{T_m}}{\argmax} \; |\mathcal{D}| \log p\left(\textcolor{unknown}{T_m}\right) + \sum_{k=1}^{|\mathcal{D}|} \log p\left(\mathbf{b}_k | \mathbf{A}_k\right). 
\end{align}
Assuming a uniform prior $p\left(\textcolor{unknown}{T_m}\right)$ over $\textcolor{unknown}{T_m}$
\begin{align}
 & \underset{\textcolor{unknown}{T_m}}{\argmax} \; p\left(\textcolor{unknown}{T_m} | \mathcal{D}\right) = \; \underset{\textcolor{unknown}{T_m}}{\argmax} \; \sum_{k=1}^{|\mathcal{D}|} \log p\left(\mathbf{b}_k | \mathbf{A}_k\right) \\
 &  = \; \underset{\textcolor{unknown}{T_m}}{\argmax} \;\sum_{k=1}^{|\mathcal{D}|} \log \mathcal{N}\left(\mathbf{b}_k; \mathbf{A}_k \mathbf{x}, \mathbf{I}\right)  \\
 &  = \; \underset{\textcolor{unknown}{T_m}}{\argmax} \; -\frac{1}{2} \sum_{k=1}^{|\mathcal{D}|} \|\mathbf{b}_k - \mathbf{A}_k \mathbf{x}\|_2^2 + C  \\
 &  = \; \underset{\textcolor{unknown}{T_m}}{\argmin} \; \sum_{k=1}^{|\mathcal{D}|} \|\mathbf{b}_k - \mathbf{A}_k \mathbf{x}\|_2^2. \label{eq:tau_optimization}
\end{align}

This has the form of a non-linear least squares problem (non-linear in $\textcolor{unknown}{T_m}$):
\begin{align}
\mathbf{A} = \begin{bmatrix}
\mathbf{A}_1 \\
\vdots \\
\mathbf{A}_{|\mathcal{D}|}]
\end{bmatrix}, \quad \mathbf{b} = \begin{bmatrix}
\mathbf{b}_1 \\
\vdots \\
\mathbf{b}_{|\mathcal{D}|}]
\end{bmatrix}. \label{eq:assembling_least_squares}
\end{align}
Since the $\argmax$ does not generally have a closed-form solution, we solve the optimization problem by a one-dimensional sweep over reasonable values of $\textcolor{unknown}{T_m}$ (bounded by the chosen support of the prior $p\left(\textcolor{unknown}{T_m}\right)$). This sweep showing the \gls*{rmse} depending on $\textcolor{unknown}{T_m}$ can be seen in Figure \ref{fig:crazyflie_finding_tau} in Section~\ref{sec:results_crazyflie}. Note that this optimization will yield $\textcolor{unknown}{T_m}$ as well as $\mathbf{x}$ which contains the parameters of the motor model/thrust-curve $\textcolor{unknown}{K_{f_{ij}}}$.

\subsection{Solving for the inertia matrix}

Based on the identified motor model we can now identify the inertia matrix. The angular dynamics are characterized by eq.~\eqref{eq:angular_dynamics} with the angular accelerations being either observable directly (e.g., in PX4) or through a finite-difference approximation based on angular velocity observations (eq.~\eqref{eq:angular_acceleration}). \gls*{wlog} we assume that the inertia matrix is aligned with the principal axes of the rigid body and hence has diagonal form: $\textcolor{unknown}{\mathbf{J}} = \diag\left(\textcolor{unknown}{I_{xx}}, \textcolor{unknown}{I_{yy}}, \textcolor{unknown}{I_{zz}}\right)$. We insert eq. \eqref{eq:torque} and obtain
\begin{align}
 & \mathbf{\textcolor{unknown}{J}} \dot{\boldsymbol{\omega}}_b - \left(\textcolor{unknown}{\mathbf{J}}\textcolor{observable}{\boldsymbol{\omega}_b}\right) \times  \textcolor{observable}{\boldsymbol{\omega}_b} = \boldsymbol{\tau} \\
 & = \; \sum_{i=1}^4 \left(\textcolor{known}{\mathbf{r}_{p_i}} \times \textcolor{known}{\mathbf{r}_{f_i}}\right)f_i + \textcolor{known}{\mathbf{r}_{\tau_i}}\textcolor{unknown}{K_{\tau_i}} f_i  \\
 & \underbrace{\mathbf{\textcolor{unknown}{J}} \dot{\boldsymbol{\omega}}_b - \left(\textcolor{unknown}{\mathbf{J}}\textcolor{observable}{\boldsymbol{\omega}_b}\right) \times  \textcolor{observable}{\boldsymbol{\omega}_b} - \sum_{i=1}^4\textcolor{known}{\mathbf{r}_{\tau_i}}\textcolor{unknown}{K_{\tau_i}} f_i}_{\mathbf{A}\mathbf{x}} = \; \underbrace{\sum_{i=1}^4 \left(\textcolor{known}{\mathbf{r}_{p_i}} \times \textcolor{known}{\mathbf{r}_{f_i}}\right)f_i}_{\mathbf{b}},  \\
\end{align}
where $f_i$ is given by eq.~\eqref{eq:thrust_curve} and the thrust curve parameters estimated in Section \ref{sec:motor_model_estimation}.
To cast this into the form $\mathbf{A}\mathbf{x} = \mathbf{b}$ we can reformulate the above terms as
\begin{align}
\left(\textcolor{unknown}{\mathbf{J}}\textcolor{observable}{\boldsymbol{\omega}_b}\right)\times  \textcolor{observable}{\boldsymbol{\omega}_b} 
= & \; \underbrace{\begin{bmatrix}
0 & \textcolor{observable}{\omega_{b_y}}\textcolor{observable}{\omega_{b_z}} & - \textcolor{observable}{\omega_{b_z}}\textcolor{observable}{\omega_{b_y}} \\
- \textcolor{observable}{\omega_{b_x}}\textcolor{observable}{\omega_{b_z}} & 0 & \textcolor{observable}{\omega_{b_z}}\textcolor{observable}{\omega_{b_x}} \\
\textcolor{observable}{\omega_{b_x}}\textcolor{observable}{\omega_{b_y}} & - \textcolor{observable}{\omega_{b_y}}\textcolor{observable}{\omega_{b_x}} & 0
\end{bmatrix}}_{\mathbf{B}} \begin{bmatrix}
\textcolor{unknown}{I_{xx}} \\
\textcolor{unknown}{I_{yy}} \\
\textcolor{unknown}{I_{zz}}
\end{bmatrix} \\
\mathbf{\textcolor{unknown}{J}} \dot{\boldsymbol{\omega}}_b = & \; \underbrace{\begin{bmatrix}
\dot{\omega}_{b_x} & 0 & 0 \\
0 & \dot{\omega}_{b_y} & 0 \\
0 & 0 & \dot{\omega}_{b_z}
\end{bmatrix}}_{\mathbf{C}} \begin{bmatrix}
\textcolor{unknown}{I_{xx}} \\
\textcolor{unknown}{I_{yy}} \\
\textcolor{unknown}{I_{zz}}
\end{bmatrix} \\
\end{align}
Boiling down to
\begin{align}
& \; \mathbf{\textcolor{unknown}{J}} \dot{\boldsymbol{\omega}}_b - \textcolor{unknown}{\mathbf{J}}\textcolor{observable}{\boldsymbol{\omega}_b} \times  \textcolor{observable}{\boldsymbol{\omega}_b} - \sum_{i=1}^4 \textcolor{known}{\mathbf{r}_{\tau_i}}\textcolor{unknown}{K_{\tau_i}} f_i = \mathbf{b} \\
& = \;  \mathbf{B} \begin{bmatrix}
\textcolor{unknown}{I_{xx}} \\
\textcolor{unknown}{I_{yy}} \\
\textcolor{unknown}{I_{zz}}
\end{bmatrix} - \mathbf{C} \begin{bmatrix}
\textcolor{unknown}{I_{xx}} \\
\textcolor{unknown}{I_{yy}} \\
\textcolor{unknown}{I_{zz}}
\end{bmatrix} - \sum_{i=1}^4 \textcolor{known}{\mathbf{r}_{\tau_i}}\textcolor{unknown}{K_{\tau_i}} f_i = \mathbf{b} \\
& = \;  \mathbf{A}_{\textcolor{unknown}{\mathbf{J}}} \begin{bmatrix}
\textcolor{unknown}{I_{xx}} \\
\textcolor{unknown}{I_{yy}} \\
\textcolor{unknown}{I_{zz}}
\end{bmatrix} + \mathbf{A}_{\textcolor{unknown}{K_{\tau}}} \begin{bmatrix}
\textcolor{unknown}{K_{\tau_1}} \\
\vdots \\
\textcolor{unknown}{K_{\tau_4}}
\end{bmatrix} = \mathbf{b}, \\
& \; \mathbf{A} \mathbf{x} = \mathbf{b},
\end{align}
with
\begin{align}
\mathbf{A}_{\textcolor{unknown}{\mathbf{J}}} := & \; \begin{bmatrix}
\dot{\omega}_{b_x} & -\textcolor{observable}{\omega_{b_y}}\textcolor{observable}{\omega_{b_z}} &  \textcolor{observable}{\omega_{b_z}}\textcolor{observable}{\omega_{b_y}}, \\
\textcolor{observable}{\omega_{b_x}}\textcolor{observable}{\omega_{b_z}} & \dot{\omega}_{b_y} & -\textcolor{observable}{\omega_{b_z}}\textcolor{observable}{\omega_{b_x}}\\
-\textcolor{observable}{\omega_{b_x}}\textcolor{observable}{\omega_{b_y}} & \textcolor{observable}{\omega_{b_y}}\textcolor{observable}{\omega_{b_x}} & \dot{\omega}_{b_z}
\end{bmatrix}, \label{eq:A_j} \\
\mathbf{A}_{\textcolor{unknown}{K_{\tau}}} := & \; 
\begin{bmatrix}
-\textcolor{known}{\mathbf{r}_{\tau_1}} f_1 & 
\ldots & 
-\textcolor{known}{\mathbf{r}_{\tau_4}} f_4
\end{bmatrix} ,\\
\textbf{A} := & \; \begin{bmatrix}
\mathbf{A}_{\textcolor{unknown}{\mathbf{J}}} & \mathbf{A}_{\textcolor{unknown}{K_{\tau}}}
\end{bmatrix}, \\
\mathbf{x} := & \; \begin{bmatrix}
\textcolor{unknown}{I_{xx}} &
\textcolor{unknown}{I_{yy}} &
\textcolor{unknown}{I_{zz}} & 
\textcolor{unknown}{K_{\tau_1}} &
\ldots &
\textcolor{unknown}{K_{\tau_4}} &
\end{bmatrix}^\top, \\
\mathbf{b} := & \; \sum_{i=1}^4 \left(\textcolor{known}{\mathbf{r}_{p_i}} \times \textcolor{known}{\mathbf{r}_{f_i}}\right)f_i
= \mathbf{R}_{pf} \mathbf{f}, \\
 \mathbf{R}_{pf} = & \; 
\begin{bmatrix}
\textcolor{known}{\mathbf{r}_{p_1}} \times \textcolor{known}{\mathbf{r}_{f_1}} & 
\ldots & 
\textcolor{known}{\mathbf{r}_{p_4}} \times \textcolor{known}{\mathbf{r}_{f_4}} 
\end{bmatrix}, \label{eq:b_inertia}\\
\mathbf{f} = & \begin{bmatrix}
f_1 & \ldots & f_4
\end{bmatrix}^\top.
\end{align}

\begin{figure*}[t]
    \centering
    \begin{minipage}{0.65\textwidth}
        \begin{adjustbox}{width=1.0\columnwidth}
        \def\arraystretch{1.2}
        \begin{tabular}{lllrrrrrr}
        \hline
         model & $\textcolor{known}{m}$ & $\textcolor{unknown}{I_{xx}}$ & $\textcolor{unknown}{I_{yy}}$ & $\textcolor{unknown}{I_{zz}}$ & $\textcolor{known}{C_{xy\rightarrow z}}$ \\
        \hline
        \grayrow
        x500 (PX4 Gazebo) & 2.000e+00 & 2.200e-02 & 2.200e-02 & 4.000e-02 & 1.818 \\
        Leshikar et al., 2021 \cite{leshikar2021asymmetric} & 2.500e+00 & 5.470e+01 & 1.560e+01 & 5.720e+01 & 1.627 \\
        \grayrow
        Kaputa et al., 2020 \cite{kaputa2020quadrotor} & 2.500e-01 & 4.270e-04 & 6.090e-04 & 1.500e-03 & 2.896 \\
        Flightmare \cite{song2021flightmare} & 7.300e-01 & 7.911e-03 & 7.911e-03 & 1.231e-02 & 1.556 \\
        \grayrow
        Iris (PX4 Gazebo) & 1.500e+00 & 2.913e-02 & 2.913e-02 & 5.523e-02 & 1.896 \\
        px4vision (PX4 Gazebo) & 1.500e+00 & 2.913e-02 & 2.913e-02 & 5.523e-02 & 1.896 \\
        \grayrow
        X-wing \cite{xu2019learning} & 1.532e+00 & 1.840e-01 & 1.910e-01 & 3.360e-01 & 1.792 \\
        Crazyflie 2.0 \cite{foerster2015systemidentificationofthecrazyflie} & 2.700e-02 & 1.660e-05 & 1.670e-05 & 2.930e-05 & 1.760 \\
        \grayrow
        Crazyflie 2.0 \cite{landry2015planning} & 2.700e-02 & 2.400e-05 & 2.400e-05 & 3.230e-05 & 1.346 \\
        Crazyflie (disk model) & 2.700e-02 & 1.389e-05 & 1.389e-05 & 2.734e-05 & 1.968 \\
        \grayrow
        Crazyflie \cite{sanca2008dynamic} & 2.700e-02 & 1.248e-05 & 1.248e-05 & 2.342e-05 & 1.876 \\
        Crazyflie \cite{luis2016design} & 2.700e-02 & 1.400e-05 & 1.400e-05 & 2.170e-05 & 1.550 \\
        \hline
        Median Crazyflie & 2.700e-2 & 1.400e-05 & 1.400e-05 & 2.734e-05 & 1.760 \\
        Mean $\textcolor{known}{C_{xy\rightarrow z}}$ & & & & & 1.832  \\
        \hline
        \end{tabular}
        \end{adjustbox}
        \vspace{-1mm}
        \captionof{table}{Inertia matrix entries of various quadrotors and the relationship between $\textcolor{unknown}{I_{xx}}$, $\textcolor{unknown}{I_{yy}}$ and $\textcolor{unknown}{I_{zz}}$ as described in Figure \ref{fig:inertia_data} (note the axes' descriptions).}
        \label{table:inertia_data}
    \end{minipage}\hfill
    \begin{minipage}{0.34\textwidth}
        \centering
        \includegraphics[width=0.95\linewidth]{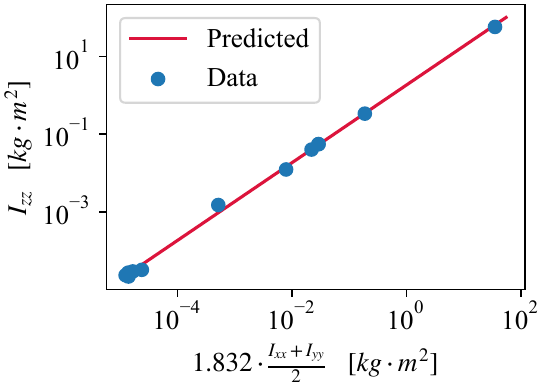}
        \vspace{-2mm}
        \caption{Data from Table \ref{table:inertia_data}.}
        \label{fig:inertia_data}
    \end{minipage}
    \vspace{-15pt}
\end{figure*}

\subsubsection{Purely vertically actuated quadrotors}
In practice, with quadrotors that have all their motors pointing straight into the $z$ direction, $\mathbf{R}_{pf}$ can have a reduced row rank. This means that e.g. the $z$ equation is homogeneous and hence can carry an ambiguity between $\textcolor{unknown}{I_{zz}}$ and $\textcolor{unknown}{K_{\tau}}$. This is especially the case if the data has been collected around the hovering point with low angular velocities. In this case, the off-diagonal terms of eq.~\eqref{eq:A_j} vanish. If there were high angular velocities, we might be able to recover the grounded value for $I_{zz}$ through the precession terms. If this is not the case we get a decoupled homogeneous system of equations of the form $\dot{\omega}_{b_z} \textcolor{unknown}{I_{zz}} - \sum_{i=1}^4 \textcolor{unknown}{K_{\tau_i}} f_i = 0$ which is only defined up to a constant factor and also has a trivial solution. 

In practice, we find that using the upper $2 \times 2$ submatrix of $\textbf{A}$ with the small angular velocity assumption to be sufficient to get a good estimate of the inertial terms around the roll and pitch axis
\begin{align}
\mathbf{A} := & \; \diag\left(\mathbf{A}_{\textcolor{unknown}{\mathbf{J}}}\right)_{1:2} = \begin{bmatrix}
\dot{\omega}_{b_x} & 0  \\
0 & \dot{\omega}_{b_y}
\end{bmatrix}, \label{eq:roll_pitch_dynamics_least_squares}\\
\mathbf{x} := & \; \begin{bmatrix}
\textcolor{unknown}{I_{xx}} &
\textcolor{unknown}{I_{yy}} &
\end{bmatrix}^\top, \label{eq:roll_pitch_dynamics_least_squares_x} \\
\textbf{b} := & \; \left(\mathbf{R}_{pf} \mathbf{f}\right)_{1:2}.
\end{align}
Since $\mathbf{A}$ and $\mathbf{b}$ are defined for each timestep we assemble them in the same way as described in eq.~\eqref{eq:assembling_least_squares} to yield a least squares problem. 

As previously described, we would still like to identify $\textcolor{unknown}{I_{zz}}$ but for purely vertically actuated quadrotors with data collected around the hovering point, the z row of each data-sample yields a homogeneous least squares problem. This means we can measure the input-output behavior of the yaw dynamics but we can only identify the ratio $\textcolor{unknown}{I_{zz}}/\textcolor{unknown}{K_{\tau}}$. Here we assume all the motors have the same $\textcolor{unknown}{K_{\tau_i}} := \textcolor{unknown}{K_{\tau}}$. Theoretically, we can estimate separate $\textcolor{unknown}{K_{\tau_i}}$ up to a common constant which would be lumped into the ratio $\textcolor{unknown}{I_{zz}}/\textcolor{unknown}{K_{\tau}}$ but practically we find a common $\textcolor{unknown}{K_{\tau}}$ to be sufficient
\begin{align}
\dot{\omega}_{b_z} \textcolor{unknown}{I_{zz}} = & \; \textcolor{unknown}{K_{\tau}} \left(\mathbf{R}_{pf} \mathbf{f}\right)_{3}, \label{eq:yaw_dynamics}  \\
\dot{\omega}_{b_z} \frac{\textcolor{unknown}{I_{zz}}}{\textcolor{unknown}{K_{\tau}}} = & \; \left(\mathbf{R}_{pf} \mathbf{f}\right)_{3}, \\
\textbf{A} = \dot{\omega}_{b_z}, \quad \textbf{x} = & \; \frac{\textcolor{unknown}{I_{zz}}}{\textcolor{unknown}{K_{\tau}}}, \quad \textbf{b} = \left(\mathbf{R}_{pf} \mathbf{f}\right)_{3}.
\end{align}
Hence we can again assemble a least squares problem according to eq.~\eqref{eq:assembling_least_squares} and find the ratio. 

Theoretically the ratio $\textcolor{unknown}{I_{zz}}/\textcolor{unknown}{K_{\tau}}$ is enough to simulate the dynamics of the drone accurately but we would like to decompose it into inertia and torque coefficient for better interpretability. Since most quadrotor-drones have similar shapes, we can establish a common linear relationship between $\textcolor{unknown}{I_{xx}}$, $\textcolor{unknown}{I_{yy}}$ and $\textcolor{unknown}{I_{zz}}$. We are able to accurately estimate the former two and would like to find a simple (e.g., linear) relationship to fix an $\textcolor{unknown}{I_{zz}}$ so that the resulting inertia tensor resembles a quadrotor. The inertia components of a particular model (constant shape, constant density and mass distribution) usually scale to the power of $5$. Using a simple example, we can show that under the previously mentioned conditions, scaling a model also just scales the inertia. For a simplified example we consider a box of dimensions $a \times b \times c$ of uniform density $\rho$ and we assume that $r>>x \; \forall x \in \{a, b, c\}$ so that we can approximate the inertia $I$ at a radius $r$ as a point mass $m$. The geometry is scaled by a factor $s$: 
\begin{align}
I = & \; m r^2, \\
m = & \; a b c \cdot \rho, \\
m(s) = & \; \left(sa\right) \left(sb\right) \left(sc\right)  \cdot \rho = s^3 \cdot abc  \cdot \rho, \\
I(s) = & \;  s^3 \cdot abc  \cdot \rho \left(s r\right)^2 = s^5 abc  \cdot \rho \cdot r^2 = s^5 I.
\end{align}
We can see that by purely scaling the geometry by $s$ (under constant shape, density and mass distribution) the inertia $I$ is also just scaled by a factor $s^5$.

Hence for a constant geometry, the ratio $\textcolor{unknown}{I_{zz}}(s)/\textcolor{unknown}{I_{xx}}(s) = s^5\textcolor{unknown}{I_{zz}}/(s^5\textcolor{unknown}{I_{xx}}) = \textcolor{unknown}{I_{zz}}/\textcolor{unknown}{I_{xx}} $ is constant across different scales. Based on this insight we collected inertia matrix values from various quadrotors in Table \ref{table:inertia_data}. From Fig. \ref{fig:inertia_data} we can see that there is a very clear linear relationship. Note the loglog scale maintains the linearity for $b=1$ (slope 1 in loglog $\Leftrightarrow$ slope 1 in the original coordinates):
\begin{align}
y = & \; (ax)^b, \\
\log(y) = & \; b\log(ax) = \log(ax).
\end{align}
Hence we can decompose the yaw dynamics from eq.~\eqref{eq:yaw_dynamics} into a new least squares problem
\begin{align}
\textcolor{known}{C_{xy\rightarrow z}} = & \; 1.832, \\
\textcolor{unknown}{I_{zz}} := & \; \frac{\textcolor{unknown}{I_{xx}}+\textcolor{unknown}{I_{yy}}}{2} \cdot \textcolor{known}{C_{xy\rightarrow z}}, \label{eq:yaw_inertia_prediction}\\
\dot{\omega}_{b_z} \frac{\textcolor{unknown}{I_{xx}}+\textcolor{unknown}{I_{yy}}}{2} \cdot \textcolor{known}{C_{xy\rightarrow z}} = & \; \textcolor{unknown}{K_{\tau}} \left(\mathbf{R}_{pf} \mathbf{f}\right)_{3}, \\
\textbf{A} = \left(\mathbf{R}_{pf} \mathbf{f}\right)_{3}, \quad \textbf{x} = & \; \textcolor{unknown}{K_{\tau}}, \quad \textbf{b} = \dot{\omega}_{b_z} \frac{\textcolor{unknown}{I_{xx}}+\textcolor{unknown}{I_{yy}}}{2} \cdot \textcolor{known}{C_{xy\rightarrow z}}.\label{eq:yaw_dynamics_torque_coeff_least_squares_problem}
\end{align}

\section{Results}

We apply our data-driven system identification method to two vastly different quadrotors: a \SI{27}{\gram} nano-quadrotor (Crazyflie 2.1) and a large \SI{3.35}{\kilo\gram} quadrotor. We chose the Crazyflie because it is a common quadrotor and multiple works have conducted system identification for it. Hence, using the Crazyflie as an example we can check if our simple method infers parameters that are plausible compared to previously estimated parameters. Additionally, we chose a large quadrotor to show that our method works across a broad range of platforms with vastly different dynamics. Using the model identified for the large quadrotor we also show how our simple system identification method can be combined with end-to-end \gls*{rl} to produce robust policies that work on large quadrotors, even under harsh, outdoor conditions.

In both cases, we find that \num{3} flights with a combined duration of about a minute are sufficient to find good estimates of the inertia and thrust-curve parameters.

\subsection{Crazyflie}
\label{sec:results_crazyflie}
For the identification of the Crazyflie's parameters, we conduct three flights exciting the linear acceleration, roll and pitch dynamics, and yaw dynamics respectively. Please refer to the supplementary video for the recording and additional analysis of these flights. 
Our data-collection method requires no additional equipment because the flights are conducted in manual flight mode, and because it only requires logging of proprioceptive measurements. 
The measurements are logged at a rate of \SI{1000}{\hertz}. Compared to e.g. using a motion capturing system and executing trajectories tailored for system identification the data can be less ``clean'' but in the following, we show that even with small amounts of data collected under suboptimal conditions we can recover good dynamics parameters.

\begin{figure}
\centering
\includegraphics[width=\linewidth]{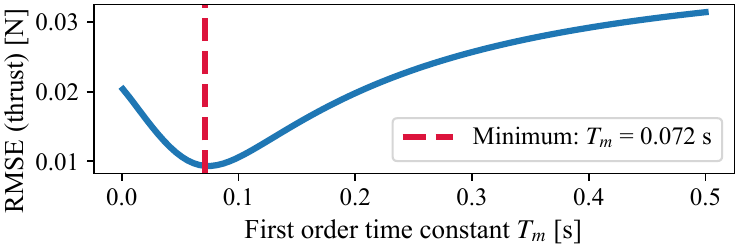}
\captionvspace{}
\vspace{-1mm}
\caption{Crazyflie: estimating $\textcolor{unknown}{T_m}$.}
\label{fig:crazyflie_finding_tau}
\vspace{-10pt}
\end{figure}

Following our method described in Section \ref{sec:methodology}, we first use the data from the first flight to find the motor delays by optimizing the objective in eq.~\eqref{eq:tau_optimization}. The search for the best $\textcolor{unknown}{T_m}$ gives rise to the curve in Fig. \ref{fig:crazyflie_finding_tau} and shows a clear optimum at $\textcolor{unknown}{T_m}=\SI{0.072}{\second}$. Based on the manufacturer's step response 
\footnote{Crazyflie motor step response: \url{https://web.archive.org/web/20220309092320/https://www.bitcraze.io/wp-content/uploads/2015/02/M1-step-response.png}}
we read the time constant to be $\approx \SI{0.073}{\second}$ (rising and falling edge averaged) which is very close to our purely data-driven estimate. Moreover, in \cite{foerster2015systemidentificationofthecrazyflie} the time constant has been estimated at \SI{0.065}{\second} which is still very close to our estimate.

\begin{figure}[h] 
\centering
\includegraphics[width=0.7\linewidth]{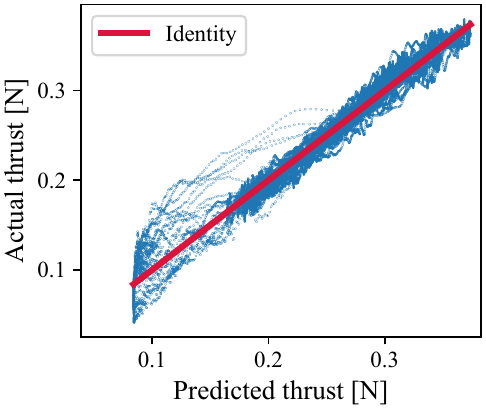}
\vspace{-2mm}
\caption{Crazyflie: resulting thrust-curve.}
\label{fig:crazyflie_thrust_curve_fit}
\vspace{-6mm}
\end{figure}

We can then use the identified $\textcolor{unknown}{T_m}$ to estimate the thrust curve. With our method (eq.~\eqref{eqn:single_sample_data_matrix}) we can identify individual thrust curves for each motor. In practice, we find that taking the mean over the found thrust curves or lumping the polynomial basis features together based on the exponents yields robust results. The result from fitting the thrust curve is shown in Fig. \ref{fig:crazyflie_thrust_curve_fit}. We can see that we are able to accurately predict observed thrusts just based on the motor command setpoints and the previously estimated motor delay. We find the best thrust curve parameters as $\textcolor{unknown}{K_{i0}}=0.0213, \textcolor{unknown}{K_{i1}}=-0.0112, \textcolor{unknown}{K_{i2}}=  0.1201$. 

 \hfill

\begin{figure}
\centering
\includegraphics[width=\linewidth]{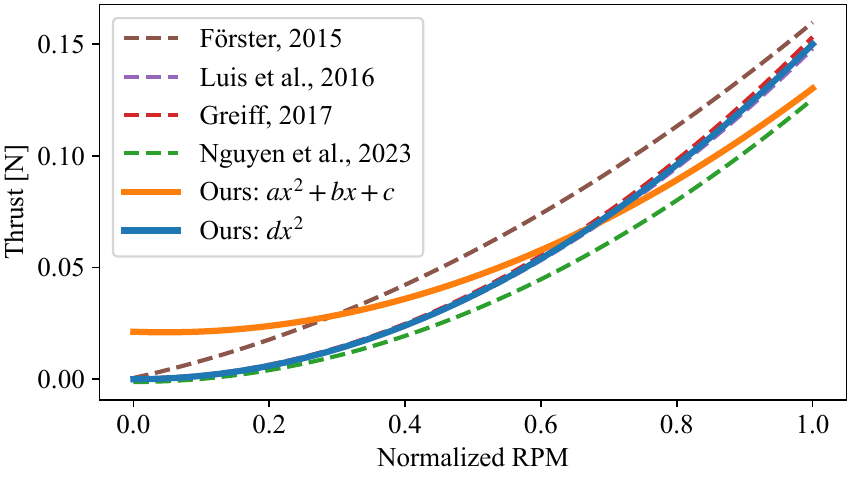}
\captionvspace{}
\caption{Crazyflie: comparison of the resulting thrust curves with \cite{foerster2015systemidentificationofthecrazyflie}, \cite{luis2016design}, \cite{greiff2017modelling}, and \cite{nguyen2023crazyflie} approaches respectively.}
\label{fig:crazyflie_thrust_curve_comparison}
\vspace{-2mm}
\end{figure}

In Fig.~\ref{fig:crazyflie_thrust_curve_comparison} we compare the thrust curves identified in the literature with our thrust curve. We also show a purely quadratic fit ($\textcolor{unknown}{K_{i2}}=0.1500$) which induces a prior on the low throttle regime ($f\left(0\right) = 0$) which is not covered well by our test flight data. From the comparison in Fig. \ref{fig:crazyflie_thrust_curve_comparison} we can see that both of our identified thrust curves are in the plausible regime. In the following, we will continue using the thrust curve with all components as we found it to predict the observed thrusts much better than the purely quadratic fit.

\begin{figure}
\centering
\includegraphics[width=\linewidth]{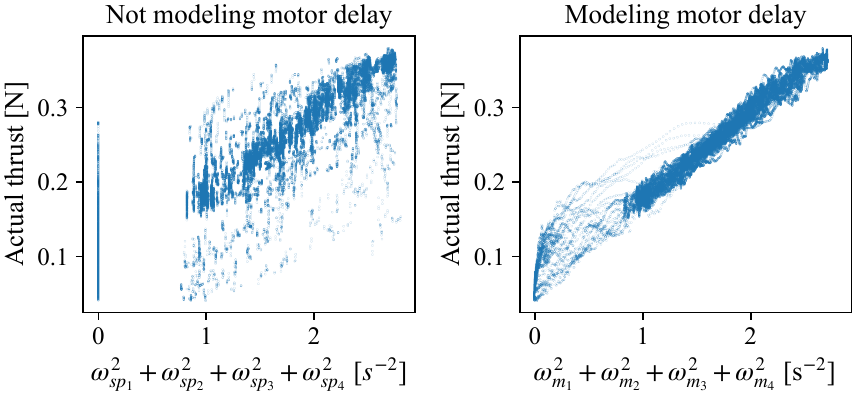}
\captionvspace{}
\caption{Crazyflie: comparing not modeling the motor delay vs modeling it (same data). 
}
\label{fig:crazyflie_thrust_curve_delay_comparison}
\vspace{-6mm}
\end{figure}

The necessity of taking into account the motor delays can be seen in Fig. \ref{fig:crazyflie_thrust_curve_delay_comparison} where the relationship between squared \gls*{rpm} setpoints and observed accelerations (Fig. \ref{fig:crazyflie_thrust_curve_delay_comparison}, left side) is very noisy and does not even appear to be linear. Whereas, when estimating and taking into account the motor delay, a strong correlation can be observed (Fig. \ref{fig:crazyflie_thrust_curve_delay_comparison}, right side). 

\begin{figure}[h]
\vspace{-1mm}
\centering
\includegraphics[width=\linewidth]{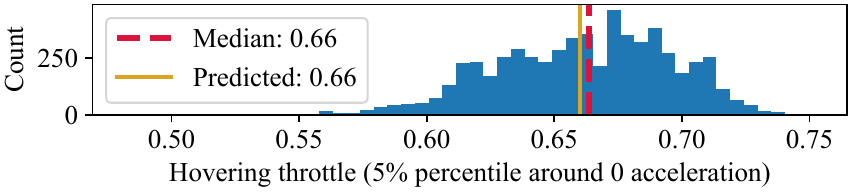}
\captionvspace{}
\caption{Crazyflie: distribution of (normalized) \glspl*{rpm} around the hovering point.}
\label{fig:crazyflie_hovering_throttle}
\vspace{-2mm}
\end{figure}

We also validate the thrust curve by investigating the distribution of $\boldsymbol{\omega}_m$ around the hover point (where $\dot{\textbf{v}} \approx 0$). We found that taking advantage of percentiles (e.g. the 5\% of the data that is around the hovering point) is a robust way to find the hovering distribution. Fig. \ref{fig:crazyflie_hovering_throttle} shows the distribution around the hovering point and shows that our thrust curve accurately predicts the hovering \glspl*{rpm}.

\begin{figure*}
\centering
\includegraphics[width=\linewidth]{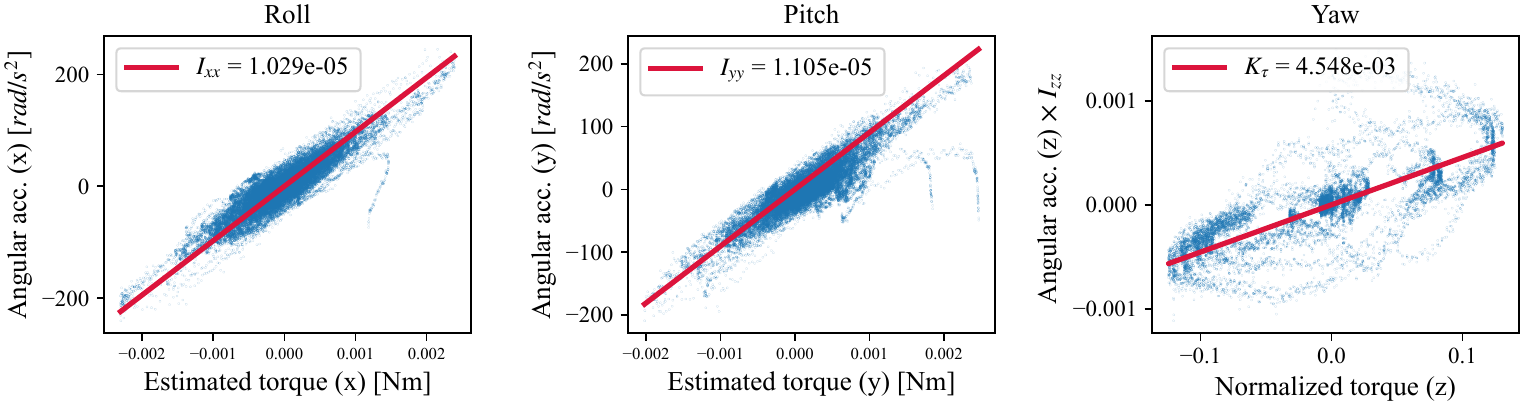}
\captionvspace{}
\vspace{-1mm}
\caption{Crazyflie: angular dynamics.}
\label{fig:crazyflie_inertia}
\vspace{-10pt}
\end{figure*}

Now we can use the estimated thrust curve to calculate the thrust of each motor at each point in time and use the method from eq.~\eqref{eq:roll_pitch_dynamics_least_squares} and following to estimate the roll and pitch inertia components. The results can be seen in Fig. \ref{fig:crazyflie_inertia}. We can see that the relationship is more noisy than in the case of the linear acceleration dynamics but we can still find a strong correlation between the estimated input torque and the angular acceleration response of the system. From a comparison of our parameters with the median across parameters reported for the Crazyflie in the literature (Table \ref{table:inertia_data}) we can see that our inertia estimates are very close (considering the high variance in the estimates in the literature). In comparison to the other approaches our method only requires a small amount of data and no complex equipment to identify the parameters.

Finally, we use the estimated inertia components $\textcolor{unknown}{I_{xx}}$ 
 and $\textcolor{unknown}{I_{yy}}$ to estimate $\textcolor{unknown}{I_{zz}}$ based on the trend observed in Table \ref{table:inertia_data} and Fig. \ref{fig:inertia_data}. Based on this we can apply eq.~\eqref{eq:yaw_inertia_prediction}. Using the found $\textcolor{unknown}{I_{zz}} = 1.955 \times 10^{-5}$ we can apply eq.~\eqref{eq:yaw_dynamics_torque_coeff_least_squares_problem} to define a least squares problem for identifying $\textcolor{unknown}{K_{\tau}}$. 
 The relationship of the yaw dynamics can be seen in Fig. \ref{fig:crazyflie_inertia}. Like in the case of the roll/pitch dynamics, the relationship is also relatively noisy. Nevertheless, our method finds $\textcolor{unknown}{K_{\tau}} = 4.548 \times 10^{-3}$ which is close to $[5.96 \times 10^{-3}, 3.73 \times 10^{-3}, 6.25 \times 10^{-3}]$ found by \cite{foerster2015systemidentificationofthecrazyflie}, \cite{landry2015planning} and \cite{greiff2017modelling} respectively. Note that we consider the estimate ``close'' because, firstly, the uncertainty about it is large (e.g., \cite{greiff2017modelling} only estimates the order of magnitude of $\textcolor{unknown}{K_{\tau}}$, without a mantissa), and secondly, the yaw dynamics are slower and less critical for stabilization in general.

\subsection{Large Quadrotor}
We apply our method from Section \ref{sec:methodology} to the large quadrotor in the same manner as for the Crazyflie before (Section \ref{sec:results_crazyflie}). 
For the large quadrotor, we again conduct three flights to excite the linear acceleration, angular roll/pitch dynamics, and the yaw dynamics respectively.

\begin{figure}[t]
\centering
\includegraphics[width=\linewidth]{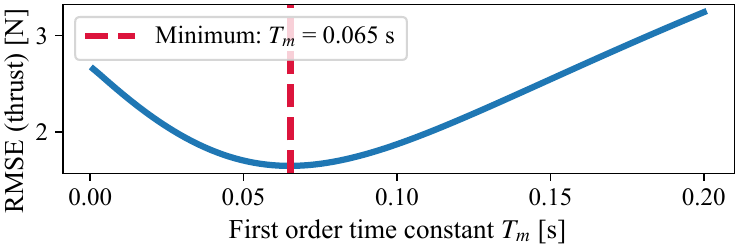}
\captionvspace{}
\vspace{-1mm}
\caption{Large quadrotor: estimating $\textcolor{unknown}{T_m}$.}
\label{fig:fs_finding_tau}
\vspace{-6mm}
\end{figure}

We use the same approach to find the motor delay of the large quadrotor (Fig. \ref{fig:fs_finding_tau}) and find that the delay is much smaller than in the case of the Crazyflie (Fig. \ref{fig:crazyflie_finding_tau}). 

\label{sec:results_large_quadrotor}

\begin{figure}
\centering
\includegraphics[width=0.7\linewidth]{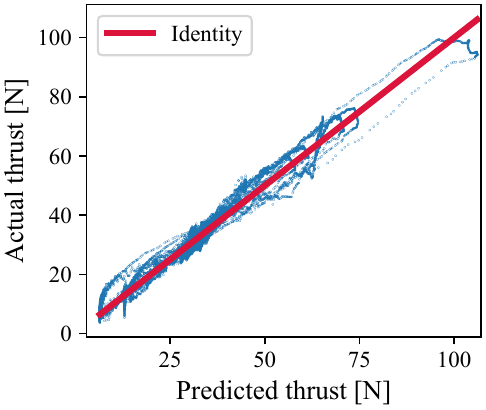}
\vspace{-2mm}
\caption{Large quadrotor: resulting thrust-curve.}
\label{fig:fs_thrust_curve_fit}
\vspace{-15pt}
\end{figure}

The resulting thrust curve for the large quadrotor shows a very good predictive performance as can be seen in Fig. \ref{fig:fs_thrust_curve_fit}. 

\begin{figure*}
\centering
\includegraphics[width=\linewidth]{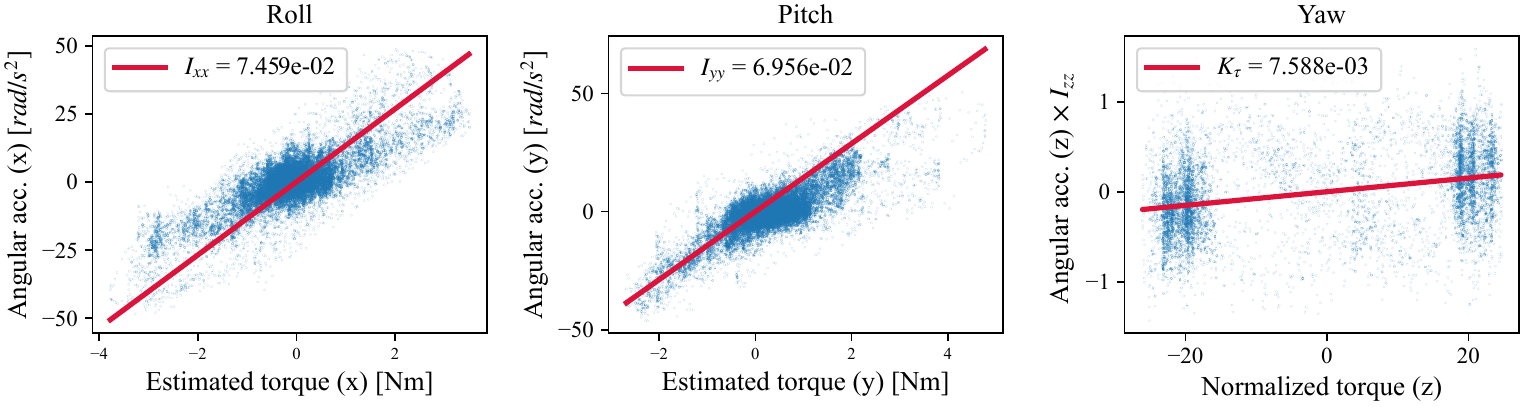}
\captionvspace{}
\vspace{-1mm}
\caption{Large quadrotor: angular dynamics.}
\label{fig:fs_inertia}
\vspace{-20pt}
\end{figure*}

The estimates for the roll and pitch inertia in Fig. \ref{fig:fs_inertia}, as before, are more noisy but still show a clear correlation. Note that especially in the case of the roll (x) inertia there appears to be another mode (with a lower slope). We find that this mode can be found by fitting the reciprocal $\textcolor{unknown}{I_{xx}}^{-1}$. Instead using the $\textcolor{unknown} {I_{xx}}$ parameterization like in eq.~\eqref{eq:roll_pitch_dynamics_least_squares} and \eqref{eq:roll_pitch_dynamics_least_squares_x}, we can formulate the reciprocal least squares problem $\mathbf{b} \mathbf{x}^{-1} = \mathbf{A}$ because the system of equations is decoupled. Here the new decision variable is $\mathbf{x}_{\text{new}} := \mathbf{x}^{-1}$ and after fitting $\mathbf{x}_{\text{new}}$ we can recover $\mathbf{x}$ by taking the reciprocal again. This finding motivates future research into the impact of the parameterization in least-squares-based system identification.

The estimation of the $\textcolor{unknown}{K_{\tau}}$ coefficient in Fig. \ref{fig:fs_inertia}, is particularly noisy but still yields a useful model as shown in the following.

\subsection{Real-world deployment}

\begin{figure}
\centering
\includegraphics[width=\linewidth]{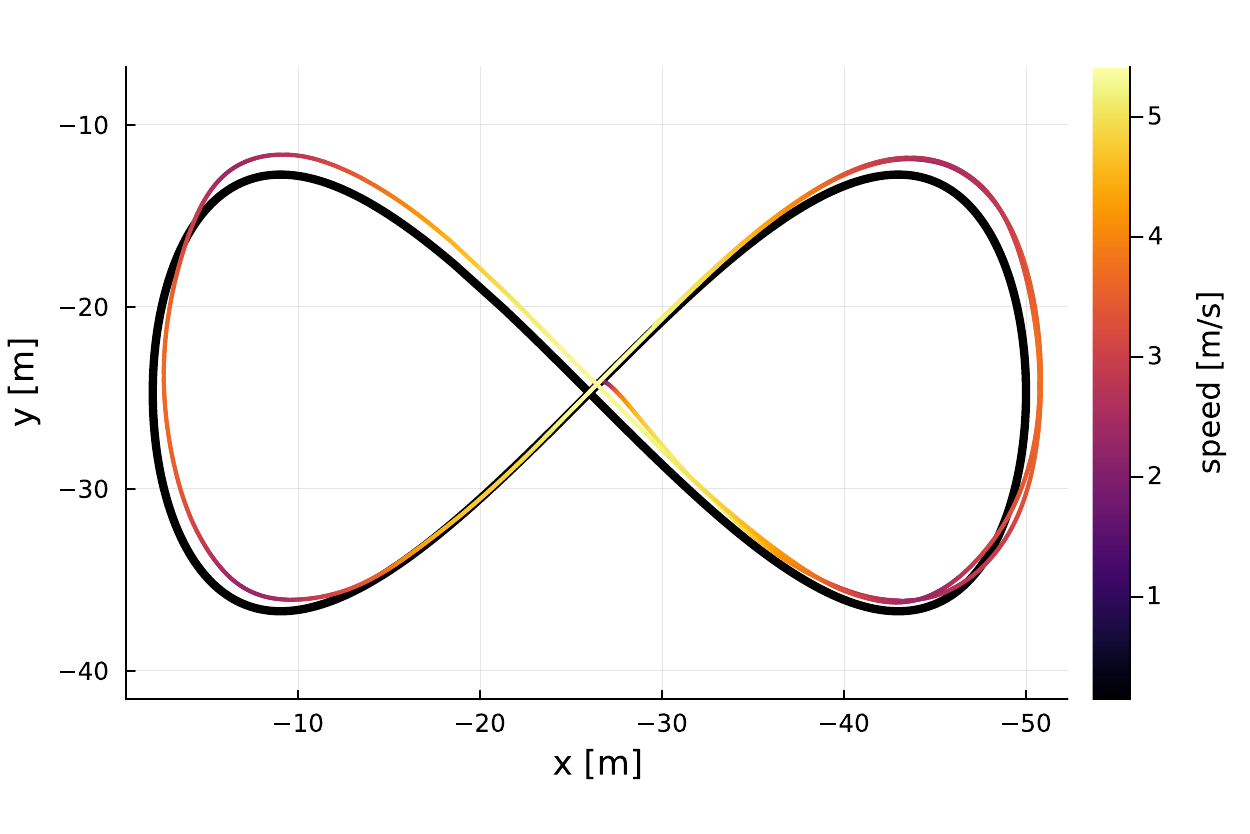}
\captionvspace{}
\vspace{-4mm}
\caption{Large quadrotor: Trajectory tracking on the real quadrotor (subject to strong wind, cf. \href{https://youtu.be/G3WGthRx2KE}{the supplementary video}).}
\label{fig:trajectory_tracking}
\vspace{-20pt}
\end{figure}

In the next step, we use the previously identified model of the large quadrotor to train an end-to-end \gls*{rl} policy that directly maps the estimated state (position, orientation, linear velocity, angular velocity, action history) to motor outputs in the form of \gls*{rpm} setpoints. We use the \gls*{rl} method introduced in \cite{eschmann2023learning} (which is built on top of \cite{eschmann2023rltools}) and find that by solely training in simulation, using the parameters identified by our method, the policy can fly the real drone even under challenging conditions like strong, gusty wind up to $8~\si{\metre\per\second}$. In Fig. \ref{fig:trajectory_tracking}, we can see the real-world trajectory tracking performance at speeds of up to $5
~\si{\metre\per\second}$. The trajectory tracking experiment was conducted under realistic, outdoor conditions as can be seen in the \href{https://youtu.be/G3WGthRx2KE}{supplementary video}.

\section{Conclusion}
In this work, we presented a simple, data-driven system identification method for quadrotors. We show that our method can recover the otherwise challenging to measure dynamics parameters of the thrust curve, torque coefficient, inertia matrix, and motor time-constant, while only requiring about a minute of flight data that can be collected with three simple maneuvers without additional equipment by just logging proprioceptive measurements. 
The main limitation is that flight data is required, but it is well known that e.g., using the default parameters in PX4 allows a drone to stay in the air (albeit probably not performing as desired, considering oscillations etc.). 
Hence, in combination with model-based methods like inverse dynamics, \gls*{mpc} or end-to-end \gls*{rl} our method can replace the tedious controller tuning which usually takes many minutes or even hours of flight time. We believe this will accelerate research by easing the adoption of new, esoteric platforms as well as paving the way for the adoption of more recent control methods in industry.

Future work will consider separate motor models for the rising and falling edge of the control inputs (relative to the current state), investigate the reciprocal parameterization mentioned in Section \ref{sec:results_large_quadrotor}, extend the system identification to more platforms, and improving the system-identification website based on user-feedback from the community.

\FloatBarrier
\bibliographystyle{IEEEtran}
\bibliography{bibliography}
\end{document}